%% file: iclr2027_conference.tex
\documentclass{article}
\usepackage{iclr2027_conference,times}

\input{math_commands.tex}

\usepackage{amsmath}
\usepackage{amssymb}
\usepackage{booktabs}
\usepackage{colortbl}
\usepackage{graphicx}
\usepackage{caption}
\usepackage{wrapfig}
\usepackage{algorithm}
\usepackage{algpseudocode}
\usepackage{hyperref}
\usepackage{url}

\title{\fontsize{15}{18}\selectfont
Calibrate the Decisions That Change the Future:\\
On-Policy Post-Training Quantization for\\
Multimodal Large Language Models}

\author{%
Wenxiao Fan\textsuperscript{1},
Jingling Fu\textsuperscript{2},
Lichen Ma\textsuperscript{2,3},
Yu He\textsuperscript{2},
Luohang Liu\textsuperscript{2},
Jinbao Xue\textsuperscript{2},\\
\textbf{Ke Zhang\textsuperscript{2},
Junshi Huang\textsuperscript{2},
Kan Li\textsuperscript{1}\thanks{Corresponding author.}}\\[0.5em]
\normalfont\small
\textsuperscript{1}School of Computer Science and Technology,
Beijing Institute of Technology\\
\normalfont\small
\textsuperscript{2}JD.com\\
\normalfont\small
\textsuperscript{3}Institute of Artificial Intelligence and Robotics,
Xi'an Jiaotong University\\[0.3em]
\normalfont\small
\texttt{wenxiaofan@bit.edu.cn}
}

\iclrfinalcopy

\newcommand{\method}{\textsc{OnPTQ}}
\newcommand{\TV}{\operatorname{TV}}
\newcommand{\JS}{\operatorname{JS}}
\definecolor{onptqrow}{RGB}{232,241,253}

\begin{document}

\maketitle

\begin{abstract}
Post-training quantization (PTQ) lowers deployment cost for multimodal large
language models, but calibration typically reconstructs fixed sequences with
local objectives. This overlooks autoregressive feedback: a quantization-induced
token change redirects the prefix and changes future states. Yet on-policy
coverage alone is insufficient because many decision mismatches barely affect
future generation. We propose \method{}, an on-policy framework that calibrates
on trajectories visited by the current quantized policy. On shared prefixes,
\method{} identifies quantization-eroded boundaries, evaluates competing tokens
through short counterfactual rollouts, and combines current discrepancy with
branch consequence into a Decision--Consequence risk. The risk prioritizes
critical states, while context anchoring and trajectory refresh preserve
multimodal behavior and keep calibration aligned with the updated policy. We
further derive a Decision--Consequence bound linking behavioral deviation to
current policy discrepancy and action-conditioned future-value span. Across
vision--language and omni-modal Qwen models under multiple low-bit settings,
\method{} improves downstream performance and yields fewer correctness flips
against the corresponding Dense/FP16 references, without changing the deployed
inference graph.
\end{abstract}

\section{Introduction}
\label{sec:introduction}

Large vision--language models (VLMs) have advanced multimodal understanding
and generation~\citep{bai2025qwen25vl,bai2025qwen3vl,xu2025qwen25omni}, but
remain costly to deploy. Post-training quantization (PTQ) reduces this cost
without full retraining through smoothing, learned transformations, rotations,
and task-aware calibration~\citep{xiao2023smoothquant,shao2024omniquant,
liu2025spinquant,sun2025flatquant,kim2025guidedquant}. VLM-specific methods
further model modality imbalance, token and channel sensitivity, and
precision allocation~\citep{li2025mbq,xiang2026qig,jia2026quantexperts,
zhang2026maba,hu2026masquant}. Despite this progress, calibration is still commonly performed on
largely fixed image--text contexts selected independently of the prefixes
produced by the quantized model. Existing objectives refine \emph{what} to
preserve at a given input, but not \emph{which autoregressive states} should
receive calibration effort.

Autoregressive generation makes this omission consequential. At step $t$, the
state $s_t=(I,x,y_{<t})$ contains the model-generated prefix; a
quantization-induced token change therefore alters the next state and all
subsequent predictions. This creates a closed-loop calibration--deployment
gap related to the train--generation mismatch studied in sequential
learning~\citep{ross2011dagger,bengio2015scheduled,ranzato2016sequence,
goyal2016professor}. Yet collecting on-policy states alone is insufficient:
many numerical discrepancies never change the trajectory. We call a state
\emph{decision-critical} when quantization erodes a competitive next-token
boundary and the competing actions lead to meaningfully different futures.
Our central question is therefore: how can PTQ follow the quantized policy
while focusing calibration on the decisions that shape future generation?

Our diagnostics expose two coupled factors: state coverage and decision
consequence. Fig.~\ref{fig:motivation} shows that dense--quantized discrepancy
amplifies after a generation fork, and that calibration on quantized-policy
states better covers deployment. However, uniformly reconstructing these
states does not by itself preserve downstream behavior, while local divergence only
weakly reflects counterfactual branch separation. Calibration must therefore
combine on-policy coverage with the future consequence of each decision.

We propose \method{} (\textbf{On}-Policy \textbf{P}ost-\textbf{T}raining
\textbf{Q}uantization), a framework that turns calibration into a
\emph{Follow--Compare--Branch--Calibrate} cycle, as illustrated in
Fig.~\ref{fig:onptq-framework}.
\method{} first follows the current quantized policy to collect the states it
actually visits. At each shared quantized prefix, it compares dense and
quantized next-token decisions to identify mismatches and near-miss boundaries,
then branches from the competing tokens through short counterfactual rollouts
to estimate how strongly the alternatives separate in the future. Boundary
erosion, same-state distribution divergence, and branch consequence together
form a Decision--Consequence risk that prioritizes critical-state calibration,
while a static context anchor preserves broader multimodal behavior. After
each update, \method{} refreshes the trajectories under the new quantized
policy, keeping the calibration distribution aligned with the model being
deployed. We further derive a Decision--Consequence bound that connects
behavioral deviation to current policy discrepancy and the span of
action-conditioned future values. All additional computation is confined to
offline calibration, leaving the deployed inference graph unchanged. Across
multiple models and quantization settings, \method{} improves downstream
performance and produces fewer correctness flips relative to the same
Dense/FP16 reference.
Our contributions are threefold:
\begin{itemize}
    \item We formulate autoregressive PTQ as an on-policy calibration problem
    and identify decision-critical states through controlled diagnostics of
    state coverage and future consequence under the quantized model's own
    generation distribution.
    \item We introduce \method{}, which combines shared-prefix decision
    comparison, short counterfactual branches, risk-weighted calibration, and
    iterative policy refresh, supported by a Decision--Consequence bound that
    separates current decision discrepancy from future consequence at each
    state along the generated trajectory.
    \item Extensive experiments across multiple VLM backbones and quantization
    settings show that \method{} consistently improves downstream performance
    while better preserving the behavior of the corresponding Dense/FP16
    models.
\end{itemize}

\section{Related Work}
\label{sec:related-work}

\noindent\textbf{Post-training quantization for large language models.}
LLM PTQ improves dense-to-low-precision fidelity through activation-outlier
handling and layer reconstruction, as in LLM.int8() and ZeroQuant, or through
second-order and activation-aware weight quantization, as in OPTQ/GPTQ and
AWQ~\citep{dettmers2022llmint8,yao2022zeroquant,frantar2023optq,lin2024awq}.
SmoothQuant and OmniQuant learn scaling or equivalent transformations from
calibration data~\citep{xiao2023smoothquant,shao2024omniquant}, while QuaRot,
SpinQuant, LRQ, FlatQuant, and GuidedQuant introduce rotations, low-rank
scaling, affine flattening, or end-loss guidance~\citep{ashkboos2024quarot,
liu2025spinquant,lee2025lrq,sun2025flatquant,kim2025guidedquant}. These methods
primarily optimize numerical fidelity on fixed calibration distributions.
\method{} instead targets autoregressive calibration by prioritizing states
visited by the current quantized policy according to their behavioral risk.

\noindent\textbf{Post-training quantization for multimodal large language models.}
Multimodal PTQ additionally models heterogeneous token statistics and
cross-modal structure. Q-VLM captures intra- and inter-block dependencies, MBQ
reweights reconstruction by modality sensitivity, and VLM-PTQ combines
asymmetric correction with channel importance~\citep{wang2024qvlm,li2025mbq,
deng2026vlmptq}. QIG, Quant Experts, MABA, and MASQuant further introduce
token-level sensitivity, adaptive compensation, hardware-aware precision
allocation, and modality-specific smoothing~\citep{xiang2026qig,
jia2026quantexperts,zhang2026maba,hu2026masquant}. This line refines how
multimodal quantization error is represented; \method{} instead addresses an
orthogonal autoregressive problem by drawing states from the current quantized
policy and prioritizing them by decision-boundary erosion and branch
consequence.

\noindent\textbf{Autoregressive distribution shift and on-policy learning.}
Autoregressive models face distribution shift because predictions determine
future inputs. DAgger, scheduled sampling, and sequence-level training expose
models to policy-induced states or their own predictions~\citep{ross2011dagger,
bengio2015scheduled,ranzato2016sequence}, while Professor Forcing aligns guided
and free-running dynamics~\citep{goyal2016professor}. ImitKD and GKD apply
related ideas to distillation on student-generated prefixes~\citep{lin2020imitkd,
agarwal2024gkd}. These methods address training or distillation; \method{}
transfers the on-policy principle to PTQ and further prioritizes visited states
by counterfactual consequence rather than weighting the rollout uniformly.

\section{Problem Setup and Motivating Observations}
\label{sec:motivation}

\noindent\textbf{Autoregressive PTQ as state-distribution matching.}
Let $D$ be a frozen dense VLM and $Q_{\theta}$ its quantized counterpart, where $\theta$ denotes the learnable quantization calibration parameters. Given an image--instruction pair $(I,x)$, the quantized model induces
\begin{equation}
    s_t=(I,x,y_{<t}), \qquad
    y_t \sim \pi_{Q_{\theta}}(\cdot\mid s_t), \qquad
    s_{t+1}=(s_t,y_t).
    \label{eq:state-transition}
\end{equation}
Here $\pi_{Q_{\theta}}$ is the next-token policy and $d_{Q_{\theta}}$ is its
induced state distribution. We denote the unlabeled calibration set of
image--instruction pairs by $\mathcal{D}_{\mathrm{cal}}$, and abbreviate
$Q_{\theta}$, $\pi_{Q_{\theta}}$, and $d_{Q_{\theta}}$ as $Q$, $\pi_Q$, and
$d_Q$ when $\theta$ is clear. Conventional reconstruction-based calibration
instead optimizes
\begin{equation}
    \min_{\theta}\;
    \mathbb{E}_{s\sim d_{\mathrm{static}}}
    \big[\mathcal{L}_{\mathrm{rec}}(D(s),Q_{\theta}(s))\big],
    \label{eq:static-ptq}
\end{equation}
where $\mathcal{L}_{\mathrm{rec}}$ is a static reconstruction objective and
$d_{\mathrm{static}}$ is the state distribution fixed by the
calibration corpus. Deployment occurs under $d_{Q_{\theta}}$, which changes
with $\theta$ and therefore turns calibration into a closed loop rather than
fixed supervised regression.

All dense--quantized comparisons use a \emph{shared quantized prefix}: $Q_{\theta}$ determines $y_{<t}$, and both models are queried at the resulting $s_t$. This isolates how quantization alters the next decision at a state the deployed model actually visits.

\begin{figure}[t]
    \centering
    \begin{minipage}[t]{0.33\textwidth}
        \vspace{0pt}
        \centering
        \makebox[\linewidth][c]{\includegraphics[height=2.85cm,keepaspectratio]{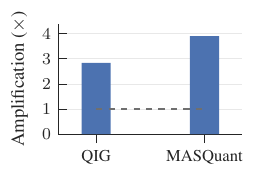}}\par
        \vspace{-1.5mm}
        {\small\textbf{(a)} Post-fork amplification}\par
    \end{minipage}\hfill%
    \begin{minipage}[t]{0.33\textwidth}
        \vspace{0pt}
        \centering
        \makebox[\linewidth][c]{\includegraphics[height=2.85cm,keepaspectratio]{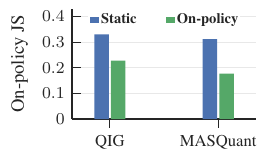}}\par
        \vspace{-1.5mm}
        {\small\textbf{(b)} On-policy coverage}\par
    \end{minipage}\hfill%
    \begin{minipage}[t]{0.33\textwidth}
        \vspace{0pt}
        \centering
        \makebox[\linewidth][c]{\includegraphics[height=2.85cm,keepaspectratio]{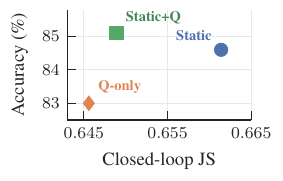}}\par
        \vspace{-1.5mm}
        {\small\textbf{(c)} Anchor trade-off}\par
    \end{minipage}
    \caption{Motivating diagnostics for \method{}. (a) Quantization discrepancy
    amplifies after a decision fork. (b) On-policy calibration better covers
    deployment states. (c) A static anchor preserves downstream accuracy.}
    \vspace{-0.4em}
    \label{fig:motivation}
\end{figure}

\noindent\textbf{Observation I: local errors are amplified after a fork.}
We compare same-state JS divergence with the divergence accumulated after dense and quantized greedy generations first choose different tokens. Under QIG and MASQuant W4A8 checkpoints, post-fork divergence is respectively $2.84\times$ and $3.91\times$ the same-state divergence (Fig.~\ref{fig:motivation}(a)), showing across two quantizers that local numerical agreement understates the discrepancy induced by a changed prefix.

\noindent\textbf{Observation II: the quantized policy visits a distinct calibration distribution.}
Fig.~\ref{fig:motivation}(b) evaluates divergence on quantized-policy states after calibration on either static or quantized-policy states. The latter reduces on-policy divergence by $30.8\%$ for QIG and $43.3\%$ for MASQuant, revealing that the deployed prefixes---not only the calibration examples---determine the relevant state distribution.

\noindent\textbf{Observation III: on-policy coverage is necessary but not sufficient.}
A naive baseline that uniformly reconstructs quantized-policy trajectories
delays the first dense--quantized fork by $3.54$ tokens, but barely changes
post-fork divergence and lowers accuracy by approximately $2.5$ points. A
static context anchor mitigates this trade-off, reducing closed-loop JS while
maintaining accuracy comparable to static calibration
(Fig.~\ref{fig:motivation}(c)). Thus, on-policy signals should complement, not
replace, the multimodal context preserved by static calibration.
Local dense--quantized JS also correlates only weakly with downstream branch
effect (Pearson $r=0.232$), so local-divergence ranking can miss subtle yet
consequential decision boundaries.

Together, these findings motivate three requirements: use states from the
current quantized policy, retain a static context anchor, and prioritize states
by current decision damage and future consequence. \method{} implements these
principles below.

\section{On-Policy Post-Training Quantization}
\label{sec:method}

\subsection{Overview}

\begin{figure}[t]
    \centering
    \includegraphics[width=\textwidth]{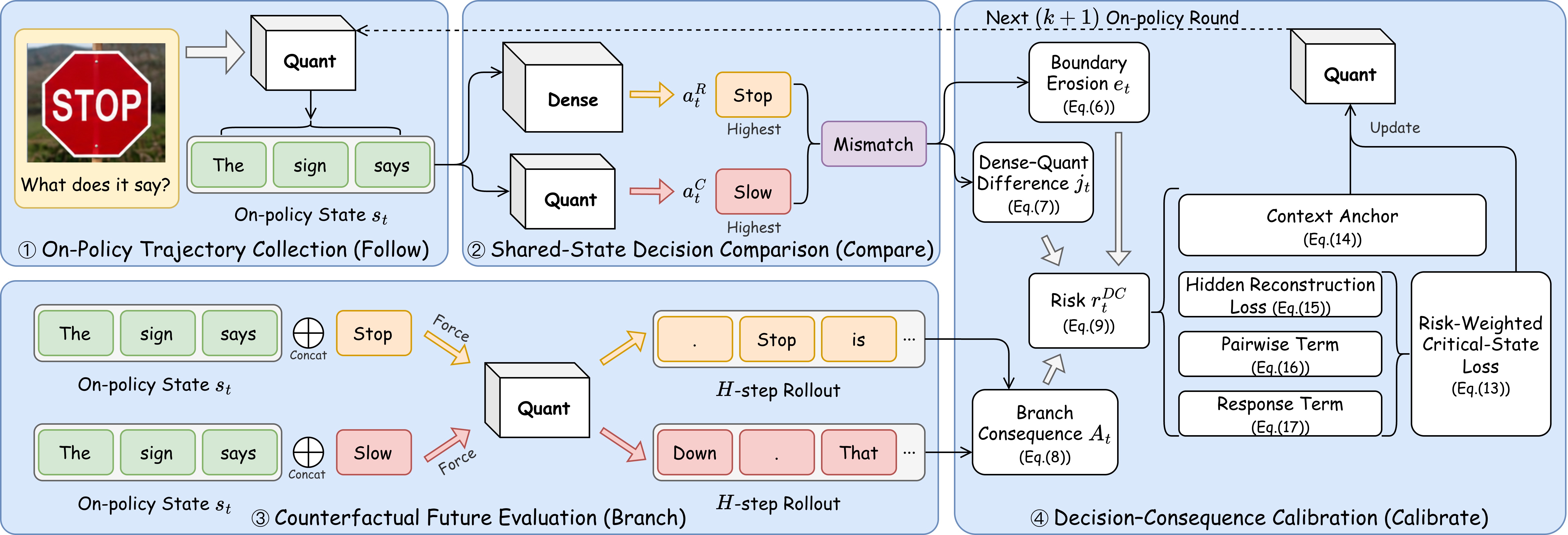}
    \caption{Overview of \method{} for a representative dense--quantized
    mismatch. \textbf{Follow} collects on-policy states, \textbf{Compare}
    identifies competing actions, \textbf{Branch} estimates future consequence,
    and \textbf{Calibrate} updates the quantizer before trajectory refresh.}
    \label{fig:onptq-framework}
\end{figure}

Fig.~\ref{fig:onptq-framework} illustrates the \method{} cycle through a representative mismatch. The current quantized model supplies calibration states (\textbf{Follow}); the dense and quantized models are compared at each shared state to identify a reference--competitor fork (\textbf{Compare}), whose short counterfactual branches estimate future consequence (\textbf{Branch}). Boundary erosion, same-state distribution divergence, and branch consequence form a Decision--Consequence risk that weights critical-state calibration, while a context anchor preserves broader multimodal behavior (\textbf{Calibrate}). Updating $\theta$ starts a new on-policy round. Secs.~\ref{sec:onptq-cycle} and~\ref{sec:policy-refresh} detail the calibration cycle and policy refresh, while Sec.~\ref{sec:theory} provides the Decision--Consequence analysis.

\subsection{The \method{} calibration cycle}
\label{sec:onptq-cycle}

\noindent\textbf{Follow: on-policy trajectory collection.}
We begin by asking \textit{which states should define calibration for the current quantized model}. In round $k$, $Q_{\theta^{(k)}}$ generates a response $y_{1:T}$ of length $T$ for each $(I,x)\in\mathcal{D}_{\mathrm{cal}}$, and every prefix induces an on-policy state $s_t=(I,x,y_{<t})$. We retain the exact generated token IDs so that subsequent dense--quantized comparisons reuse the same prefixes without retokenization drift. These states form the on-policy trajectory bank
\[
    \mathcal{B}^{(k)}
    =\left\{s_t=(I,x,y_{<t})
      \;\middle|\;
      (I,x)\in\mathcal{D}_{\mathrm{cal}},\;
      y_{1:T}\sim\pi_{Q_{\theta^{(k)}}},\;1\leq t\leq T
      \right\},
\]
which samples $d_{Q_{\theta^{(k)}}}$ and provides the shared states used by \textbf{Compare}.

\noindent\textbf{Compare: shared-prefix decision comparison.}
We next ask \textit{which next-token decisions have changed ordering, or are close to doing so, at states visited by the current quantized model}. Let $\mathcal{A}$ denote the vocabulary. At each $s_t\in\mathcal{B}^{(k)}$, we feed the same quantized on-policy state to both models and obtain the dense and quantized next-token logits $z_D^t$ and $z_Q^t$ over $\mathcal{A}$. We then define
\begin{equation}
    a_t^R = \arg\max_a z_D^t(a),
    \qquad
    \hat a_t = \arg\max_a z_Q^t(a).
\end{equation}
We define the competing action as
\begin{equation}
    a_t^C=\arg\max_{a\neq a_t^R}z_Q^t(a).
\end{equation}
The dense and quantized pairwise margins are
\begin{equation}
    m_D^t=z_D^t(a_t^R)-z_D^t(a_t^C), \qquad
    m_Q^t=z_Q^t(a_t^R)-z_Q^t(a_t^C),
    \label{eq:pairwise-margins}
\end{equation}
with boundary erosion
\begin{equation}
    e_t=[m_D^t-m_Q^t]_+.
    \label{eq:erosion}
\end{equation}
If $\hat a_t\neq a_t^R$, $s_t$ is a \emph{mismatch} and $a_t^C=\hat a_t$; if $\hat a_t=a_t^R$ but $e_t>0$, it is a \emph{near miss}: the top action is preserved while its margin over the strongest quantized runner-up has been eroded. We additionally compute
\begin{equation}
    j_t=\JS\!\left(\pi_D(\cdot\mid s_t),
                     \pi_Q(\cdot\mid s_t)\right).
    \label{eq:same-state-js}
\end{equation} 
where $\JS$ is the Jensen--Shannon divergence. Mismatches and near misses form the candidate set $\mathcal{C}_{\theta^{(k)}}\subseteq\mathcal{B}^{(k)}$. Here, $R$ denotes the token preferred by the dense reference model, whereas $C$ denotes its strongest competing token under the current quantized model. These superscripts describe the tokens' relative roles in the decision comparison; they neither require task labels nor imply that either token is task-correct. We omit the round index when it is clear from context.

\noindent\textbf{Branch: counterfactual consequence and risk.}
Among the candidate forks, we next ask \textit{which quantization-affected next-token decisions lead to substantially different futures}. At each candidate state, we force $a_t^R$ and $a_t^C$, then let the current quantized model greedily continue both branches over a small fixed horizon $H$. We use $s_{t+h}^{R}$ and $s_{t+h}^{C}$ to denote the resulting branch states at offset $h$. The dense model evaluates their predictive separation:
\begin{equation}
    A_t=\frac{1}{H_t}\sum_{h=1}^{H_t}
    \JS\!\left(\pi_D(\cdot\mid s_{t+h}^{R}),
                 \pi_D(\cdot\mid s_{t+h}^{C})\right),
    \label{eq:amplification}
\end{equation}
where $H_t\leq H$ is the common valid horizon under early termination. Thus, $Q$ generates both futures, while $D$ measures whether they are behaviorally distinct.

We define the Decision--Consequence risk of state $s_t$ as
\begin{equation}
    {r_t^{\mathrm{DC}}
    =e_t\big(1+\lambda_b A_t\big)+\eta j_t.}
    \label{eq:risk}
\end{equation}
Here $\lambda_b$ and $\eta$ balance the branch and distribution terms: $A_t$ amplifies an eroded decision boundary, while $j_t$ provides a full-vocabulary backstop. Given a critical-state budget $K_c$, \method{} selects the $K_c$ states with the highest risk:
\begin{equation}
    \mathcal{S}_{\mathrm{critical}}
    =\operatorname{TopK}_{s_t\in\mathcal{C}_{\theta}}
      \!\left(r_t^{\mathrm{DC}}\right).
    \label{eq:critical-set}
\end{equation}
The selected risks are normalized as
\begin{equation}
    \widetilde w_t=
    \frac{r_t^{\mathrm{DC}}}
    {\sum_{s_u\in\mathcal{S}_{\mathrm{critical}}}r_u^{\mathrm{DC}}}.
    \label{eq:normalized-risk}
\end{equation}

\begin{table*}[t]
    \centering
    \caption{Accuracy (\%, $\uparrow$) across precision settings. OCR:
    OCRBench; Viz: VizWiz; S-QA: \mbox{ScienceQA}; T-VQA: TextVQA. Avg: unweighted
    mean over the five benchmarks.}
    \label{tab:main-results}
    \setlength{\tabcolsep}{2.2pt}
    \renewcommand{\arraystretch}{0.92}
    \resizebox{\textwidth}{!}{%
    \begin{tabular}{lllcccccc@{\hspace{7pt}}cccccc}
        \toprule
        & & & \multicolumn{6}{c}{\textbf{Qwen2.5-VL-7B}} & \multicolumn{6}{c}{\textbf{Qwen3-VL-8B-Instruct}} \\
        \cmidrule(lr){4-9}\cmidrule(lr){10-15}
        \textbf{Method} & \textbf{Venue} & \textbf{Bits}
        & \textbf{MMMU} & \textbf{OCR} & \textbf{Viz} & \textbf{S-QA} & \textbf{T-VQA} & \textbf{Avg}
        & \textbf{MMMU} & \textbf{OCR} & \textbf{Viz} & \textbf{S-QA} & \textbf{T-VQA} & \textbf{Avg} \\
        \midrule
        Dense & -- & FP16
        & 46.7 & 83.8 & 70.8 & 88.4 & 82.9 & 74.5
        & 51.6 & 86.2 & 69.3 & 94.6 & 81.6 & 76.6 \\
        \midrule
        RTN & -- & W4A16
        & 43.3 & 83.7 & 67.8 & 81.3 & 82.1 & 71.6
        & 51.6 & 75.8 & 70.5 & 92.5 & 80.3 & 74.1 \\
        SmoothQuant & ICML'23 & W4A16
        & 40.7 & 79.4 & 67.3 & 81.9 & 81.7 & 70.2
        & 49.0 & 71.5 & 70.0 & 93.1 & 79.9 & 72.7 \\
        MBQ & CVPR'25 & W4A16
        & 44.4 & 82.8 & 70.6 & 87.8 & \textbf{82.9} & 73.7
        & 50.6 & 73.2 & 70.0 & 93.8 & 80.8 & 73.7 \\
        QIG & CVPR'26 & W4A16
        & 48.2 & 82.6 & 70.5 & \textbf{88.0} & 82.5 & \textbf{74.4}
        & 51.8 & 74.0 & 69.7 & 93.5 & 80.1 & 73.8 \\
        MASQuant & CVPR'26 & W4A16
        & 44.4 & \textbf{84.6} & \textbf{71.5} & 87.8 & 82.5 & 74.2
        & 49.4 & \textbf{85.2} & 69.2 & 92.9 & \textbf{81.3} & 75.6 \\
        \rowcolor{onptqrow}
        \method{} & Ours & W4A16
        & \textbf{48.3} & 83.2 & 70.1 & 87.1 & 81.9 & 74.1
        & \textbf{52.0} & 85.1 & 70.7 & 94.3 & 81.1 & \textbf{76.6} \\
        \midrule
        RTN & -- & W8A8
        & 45.6 & 83.8 & 70.5 & 88.1 & 82.5 & 74.1
        & 50.8 & 75.1 & 67.8 & 94.1 & 80.1 & 73.6 \\
        SmoothQuant & ICML'23 & W8A8
        & 43.3 & 83.8 & 70.0 & 88.2 & 82.6 & 73.6
        & 51.4 & 75.8 & 67.9 & \textbf{94.7} & 80.2 & 74.0 \\
        MBQ & CVPR'25 & W8A8
        & 46.7 & 83.5 & 70.6 & 88.5 & \textbf{82.9} & 74.4
        & 50.3 & 75.7 & 68.6 & 94.2 & 80.3 & 73.8 \\
        QIG & CVPR'26 & W8A8
        & 48.8 & 83.3 & \textbf{71.1} & 87.6 & 82.6 & 74.7
        & 51.0 & 75.1 & 68.8 & 94.3 & 80.5 & 73.9 \\
        MASQuant & CVPR'26 & W8A8
        & 46.2 & \textbf{84.2} & 70.6 & \textbf{88.6} & 82.6 & 74.4
        & \textbf{52.0} & \textbf{86.0} & 69.0 & 94.2 & \textbf{81.3} & \textbf{76.5} \\
        \rowcolor{onptqrow}
        \method{} & Ours & W8A8
        & \textbf{50.3} & 83.2 & 70.3 & 87.8 & 82.8 & \textbf{74.9}
        & 51.7 & 85.6 & \textbf{69.9} & 93.9 & 80.6 & 76.3 \\
        \midrule
        RTN & -- & W4A8
        & 43.3 & 68.3 & 63.2 & 85.2 & 76.9 & 67.4
        & 50.6 & 76.3 & 70.0 & 92.2 & 79.6 & 73.8 \\
        SmoothQuant & ICML'23 & W4A8
        & 37.8 & 70.2 & 61.5 & 83.3 & 71.1 & 64.8
        & 49.1 & 70.7 & 69.6 & 92.8 & 79.6 & 72.4 \\
        MBQ & CVPR'25 & W4A8
        & 43.3 & 74.1 & 64.3 & \textbf{86.0} & 74.8 & 68.5
        & \textbf{51.8} & 74.5 & 69.0 & 92.5 & 80.6 & 73.7 \\
        QIG & CVPR'26 & W4A8
        & 40.4 & 64.9 & 58.6 & 83.2 & 73.5 & 64.1
        & 50.6 & 75.2 & 70.6 & \textbf{93.0} & 80.1 & 73.9 \\
        MASQuant & CVPR'26 & W4A8
        & 43.3 & 72.8 & \textbf{66.4} & 85.7 & 77.0 & 69.0
        & 51.2 & \textbf{85.0} & 69.5 & 92.1 & 80.8 & 75.7 \\
        \rowcolor{onptqrow}
        \method{} & Ours & W4A8
        & \textbf{44.8} & \textbf{76.7} & 64.8 & 85.8 & \textbf{77.3} & \textbf{69.9}
        & 49.1 & 84.8 & \textbf{71.0} & \textbf{93.0} & \textbf{81.4} & \textbf{75.9} \\
        \midrule
        RTN & -- & W4A6
        & 23.7 & 52.9 & 54.7 & 27.2 & 66.9 & 45.1
        & 31.0 & 60.9 & 61.5 & 34.2 & 69.6 & 51.5 \\
        SmoothQuant & ICML'23 & W4A6
        & 28.9 & 67.7 & 60.5 & 78.6 & 69.3 & 61.0
        & 46.2 & 66.6 & \textbf{67.8} & 88.0 & 77.1 & 69.1 \\
        MBQ & CVPR'25 & W4A6
        & 30.0 & 71.7 & 59.8 & 80.1 & 72.9 & 62.9
        & 45.7 & 74.4 & 67.6 & 88.9 & \textbf{78.9} & 71.1 \\
        QIG & CVPR'26 & W4A6
        & 35.4 & 62.7 & 55.4 & 68.4 & 68.4 & 58.1
        & 47.0 & 72.5 & 65.6 & 88.1 & 78.3 & 70.3 \\
        MASQuant & CVPR'26 & W4A6
        & 29.7 & 70.3 & \textbf{62.6} & 79.7 & 72.9 & 63.0
        & 44.1 & 82.7 & 66.8 & 87.2 & 77.3 & 71.6 \\
        \rowcolor{onptqrow}
        \method{} & Ours & W4A6
        & \textbf{37.7} & \textbf{72.8} & 61.8 & \textbf{81.9} & \textbf{74.9} & \textbf{65.8}
        & \textbf{48.8} & \textbf{83.0} & 67.2 & \textbf{89.7} & 78.2 & \textbf{73.4} \\
        \bottomrule
    \end{tabular}%
    }
        \vspace{-0.6em}

\end{table*}

\noindent\textbf{Calibrate: branch-aware calibration objective.}
Having identified the critical decision states, we ask \textit{how to repair their quantization-eroded boundaries without disrupting the model's broader multimodal behavior}. \method{} separates these two roles: a context anchor preserves representations over the original multimodal context, while a risk-weighted loss concentrates calibration on selected on-policy decision states:
\begin{equation}
    \mathcal{L}_{\mathrm{OnPTQ}}
    =\mathcal{L}_{\mathrm{anchor}}
     +\gamma\sum_{s_t\in\mathcal{S}_{\mathrm{critical}}}
       \widetilde w_t\,\mathcal{L}_{\mathrm{critical}}(s_t),
    \label{eq:generic-objective}
\end{equation}
\begin{equation}
    \mathcal{L}_{\mathrm{critical}}(s_t)
    =\mathcal{L}_{\mathrm{hidden}}(s_t)
     +\mathcal{L}_{\mathrm{pair}}(s_t)
     +\mathcal{L}_{\mathrm{response}}(s_t).
    \label{eq:critical-loss}
\end{equation}
Here $\gamma$ controls the overall contribution of critical-state calibration.
For $M\in\{D,Q\}$, let $\mathcal{I}_{\mathrm{blk}}$ be the set of calibrated
blocks and let $h_{M,\ell}^{u}\in\mathbb{R}^{d_\ell}$ be the hidden vector at
token $u$ after block $\ell\in\mathcal{I}_{\mathrm{blk}}$, where $d_\ell$ is
the corresponding hidden width. Let $\mathcal{U}$ contain the image,
instruction, and prompt-prefix anchor positions. With
$\operatorname{MSE}(p,q)=\|p-q\|_2^2/d_\ell$,
\begin{equation}
    \mathcal{L}_{\mathrm{anchor}}
    =\frac{1}{|\mathcal{U}|\,|\mathcal{I}_{\mathrm{blk}}|}
      \sum_{u\in\mathcal{U}}\sum_{\ell\in\mathcal{I}_{\mathrm{blk}}}
      \operatorname{MSE}\!\left(h_{D,\ell}^{u},h_{Q,\ell}^{u}\right).
    \label{eq:context-loss}
\end{equation}
At critical state $s_t$, let $h_{M,\ell}(s_t)$ be the final-prefix hidden vector. Its reconstruction loss is
\begin{equation}
    \mathcal{L}_{\mathrm{hidden}}(s_t)
    =\frac{1}{|\mathcal{I}_{\mathrm{blk}}|}\sum_{\ell\in\mathcal{I}_{\mathrm{blk}}}
      \operatorname{MSE}\!\left(
        h_{D,\ell}(s_t),h_{Q,\ell}(s_t)\right).
    \label{eq:hidden-loss}
\end{equation}

For $\Delta m_t=m_Q^t-m_D^t$, the pairwise term uses the unit-threshold Huber loss~\citep{huber1964robust}:
\begin{equation}
    \mathcal{L}_{\mathrm{pair}}(s_t)
    =\operatorname{Huber}(\Delta m_t)
    =\begin{cases}
        \frac{1}{2}(\Delta m_t)^2, & |\Delta m_t|\leq 1,\\
        |\Delta m_t|-\frac{1}{2}, & |\Delta m_t|>1.
      \end{cases}
    \label{eq:margin-loss}
\end{equation}
The response term preserves the full next-token distribution:
\begin{equation}
    \mathcal{L}_{\mathrm{response}}(s_t)
    =\mathrm{KL}\!\left(\pi_D(\cdot\mid s_t)\,\|\,
                          \pi_Q(\cdot\mid s_t)\right).
    \label{eq:response-loss}
\end{equation}
The anchor prevents broad context drift; the hidden and response terms stabilize each critical state; and the pairwise term restores the reference--competitor boundary without reconstructing every logit equally. Calibration requires neither task labels nor a reward model.

\subsection{Iterative policy refresh}
\label{sec:policy-refresh}

Because calibration changes both the quantized policy and its induced state distribution, $\mathcal{B}^{(k)}$ becomes off-policy once $\theta^{(k)}$ is updated to $\theta^{(k+1)}$. \method{} therefore refreshes the trajectory bank in every round:
\begin{equation}
    Q_{\theta^{(k)}}\xrightarrow{\mathrm{Follow}}\mathcal{B}^{(k)}
    \xrightarrow{\mathrm{Compare,\,Branch,\,Calibrate}}
    Q_{\theta^{(k+1)}}.
\end{equation}
The next round then recollects $\mathcal{B}^{(k+1)}$ under $Q_{\theta^{(k+1)}}$. Thus, the current quantized model always determines the visited prefixes and branch continuations, while the dense model remains a frozen reference.

\begin{table*}[t]
    \centering
    \caption{Results on Qwen2.5-Omni under W4A8 and W4A6. Libri/Wen: WER
    (\%, $\downarrow$) on \mbox{LibriSpeech}/\mbox{WenetSpeech}; MMMU/Omni: accuracy
    (\%, $\uparrow$) on MMMU/OmniBench.}
    \label{tab:omni-results}
    \setlength{\tabcolsep}{4.2pt}
    \renewcommand{\arraystretch}{0.98}
    \resizebox{\textwidth}{!}{%
    \begin{tabular}{lllcccc@{\hspace{10pt}}cccc}
        \toprule
        & & & \multicolumn{4}{c}{\textbf{Qwen2.5-Omni-3B}}
        & \multicolumn{4}{c}{\textbf{Qwen2.5-Omni-7B}} \\
        \cmidrule(lr){4-7}\cmidrule(lr){8-11}
        & & & \multicolumn{2}{c}{\textbf{Audio--Text}}
        & \textbf{Vision--Text} & \textbf{Omni-modal}
        & \multicolumn{2}{c}{\textbf{Audio--Text}}
        & \textbf{Vision--Text} & \textbf{Omni-modal} \\
        \cmidrule(lr){4-5}\cmidrule(lr){6-6}\cmidrule(lr){7-7}
        \cmidrule(lr){8-9}\cmidrule(lr){10-10}\cmidrule(lr){11-11}
        \textbf{Method} & \textbf{Venue} & \textbf{Bits}
        & \textbf{Libri$\downarrow$} & \textbf{Wen$\downarrow$}
        & \textbf{MMMU$\uparrow$} & \textbf{Omni$\uparrow$}
        & \textbf{Libri$\downarrow$} & \textbf{Wen$\downarrow$}
        & \textbf{MMMU$\uparrow$} & \textbf{Omni$\uparrow$} \\
        \midrule
        Dense & -- & FP16
        & 3.9 & 7.5 & 43.3 & 43.8 & 2.9 & 7.1 & 50.0 & 45.3 \\
        \midrule
        RTN & -- & W4A8
        & 109.7 & 105.6 & 28.9 & 29.7 & 9.0 & 8.7 & 42.2 & 35.2 \\
        SmoothQuant & ICML'23 & W4A8
        & 77.4 & 94.2 & 30.0 & 27.3 & 8.6 & 8.3 & 42.2 & 36.7 \\
        MBQ & CVPR'25 & W4A8
        & 9.5 & 8.5 & 27.8 & 36.7 & 3.8 & 8.2 & 47.8 & 40.6 \\
        QIG & CVPR'26 & W4A8
        & 8.3 & 8.5 & 28.8 & 37.0 & 3.6 & 8.2 & 41.9 & 41.1 \\
        MASQuant & CVPR'26 & W4A8
        & 3.6 & 8.7 & \textbf{36.7} & 41.4
        & 2.9 & 8.0 & \textbf{48.8} & 43.8 \\
        \rowcolor{onptqrow}
        \method{} & Ours & W4A8
        & \textbf{3.5} & \textbf{8.4} & 36.4 & \textbf{42.0}
        & \textbf{2.8} & \textbf{7.8} & 43.3 & \textbf{44.0} \\
        \midrule
        RTN & -- & W4A6
        & 87.8 & 99.6 & 28.9 & 23.4 & 11.8 & 15.7 & 28.9 & 40.6 \\
        SmoothQuant & ICML'23 & W4A6
        & 87.8 & 99.5 & 28.9 & 18.8 & 5.8 & 19.8 & 37.8 & 35.9 \\
        MBQ & CVPR'25 & W4A6
        & 10.8 & 10.4 & 31.4 & \textbf{46.9} & 6.4 & 14.8 & 33.3 & 42.1 \\
        QIG & CVPR'26 & W4A6
        & 9.4 & 10.1 & 34.8 & 44.1 & 6.1 & 13.6 & 36.1 & 38.0 \\
        MASQuant & CVPR'26 & W4A6
        & 3.7 & 8.9 & 32.2 & 42.2
        & 4.7 & 8.7 & 36.8 & 42.2 \\
        \rowcolor{onptqrow}
        \method{} & Ours & W4A6
        & \textbf{3.5} & \textbf{8.5} & \textbf{35.4} & 44.5
        & \textbf{4.5} & \textbf{8.3} & \textbf{38.1} & \textbf{43.4} \\
        \bottomrule
    \end{tabular}%
    }
    \vspace{-0.6em}
\end{table*}

\subsection{Decision--Consequence Analysis}
\label{sec:theory}

We formalize why behavioral risk depends jointly on current decision discrepancy and future consequence. For a generated trajectory $\tau$, fixed continuation policy $\mu$, and bounded future utility $G(\tau)$, define
\begin{align}
    V_{\mu}(s,a)
    &=\mathbb{E}\!\left[G(\tau)\mid s,a,
      \tau_{>t}\sim\mu\right],\\
    J_{M\rightarrow\mu}(s)
    &=\mathbb{E}_{a\sim\pi_M(\cdot\mid s)}[V_{\mu}(s,a)],
    \qquad M\in\{D,Q\}.
\end{align}

\paragraph{Theorem 1 (Decision--Consequence bound).}
At any shared state $s$,
\begin{equation}
    \left|J_{Q\rightarrow\mu}(s)-J_{D\rightarrow\mu}(s)\right|
    \leq
    \underbrace{\TV\!\left(\pi_Q(\cdot\mid s),\pi_D(\cdot\mid s)\right)}_{\text{Decision}}
    \underbrace{\operatorname{span}_{a}V_{\mu}(s,\cdot)}_{\text{Consequence}},
    \label{eq:dc-bound}
\end{equation}
where $\TV(p,q)=\frac{1}{2}\sum_a|p(a)-q(a)|$ and $\operatorname{span}_{a}V_{\mu}=\max_aV_{\mu}(s,a)-\min_aV_{\mu}(s,a)$. The bound shows that either factor vanishes when the current policies agree or all actions have the same future value. It motivates, rather than certifies, Equation~\ref{eq:risk}: $e_t$ and $j_t$ proxy decision discrepancy, while $A_t$ proxies future consequence. The proof and two-action specialization appear in Appendix~\ref{app:derivations}.

\section{Experiments}
\label{sec:experiments}

\subsection{Experimental Setups}
\label{sec:experimental-setups}

\noindent\textbf{Models and quantization.}
We evaluate \method{} on Qwen2.5-VL-7B, Qwen3-VL-8B-Instruct, and the
3B/7B Qwen2.5-Omni models~\citep{bai2025qwen25vl,bai2025qwen3vl,xu2025qwen25omni}.
Following prior multimodal PTQ protocols~\citep{xiang2026qig,hu2026masquant},
we quantize the language-model
component while retaining the modality encoders at their original precision.
We consider W4A16, W8A8, W4A8, and W4A6 for vision--language models, and the
two settings, W4A8 and W4A6, for omni-modal models. Here,
W$b_w$A$b_a$ denotes $b_w$-bit weights and $b_a$-bit activations.

\noindent\textbf{Calibration datasets.}
Following QIG~\citep{xiang2026qig}, all matched vision--language methods share
a fixed manifest of 128 image--caption pairs randomly drawn from the
ShareGPT4V-improved COCO Caption dataset. \method{} uses these common prompts
as initial contexts and collects calibration states by rolling out the current
quantized model. 

\noindent\textbf{Benchmarks and metrics.}
We compare against RTN, AWQ~\citep{lin2024awq},
SmoothQuant~\citep{xiao2023smoothquant}, MBQ~\citep{li2025mbq},
QIG~\citep{xiang2026qig}, and MASQuant~\citep{hu2026masquant}. 
Vision--language performance is evaluated on MMMU, OCRBench, VizWiz,
ScienceQA, and TextVQA using task accuracy~\citep{yue2024mmmu,liu2024ocrbench,
gurari2018vizwiz,lu2022scienceqa,singh2019textvqa}. For Qwen2.5-Omni, we
additionally report word error rate on LibriSpeech and WenetSpeech and accuracy
on OmniBench~\citep{panayotov2015librispeech,zhang2022wenetspeech,
li2024omnibench}. The vision--language average is the unweighted mean over its
five benchmarks. We also report sample-wise Flip Rate against the same Dense/FP16
reference for the closed-choice MMMU and ScienceQA-IMG evaluations. Full
calibration, decoding, provenance, and evaluation details are provided in
Appendix~\ref{app:implementation-details}; hyperparameter settings and
sensitivity are reported in Appendix~\ref{app:hyperparameters}.

\subsection{Main results}
\label{sec:main-results}

\noindent\textbf{Vision-Language MLLMs.}
Tab.~\ref{tab:main-results} compares Dense/FP16 and four quantization
settings on Qwen2.5-VL-7B and
Qwen3-VL-8B-Instruct~\citep{bai2025qwen25vl,bai2025qwen3vl}.
\method{} achieves the highest average accuracy in six 
settings and leads on both backbones under W4A6. Its
advantage is most consistent at lower activation precision, while remaining
competitive under W4A16 and W8A8.

\begin{figure}[htbp]
    \centering
    \includegraphics[width=\textwidth]{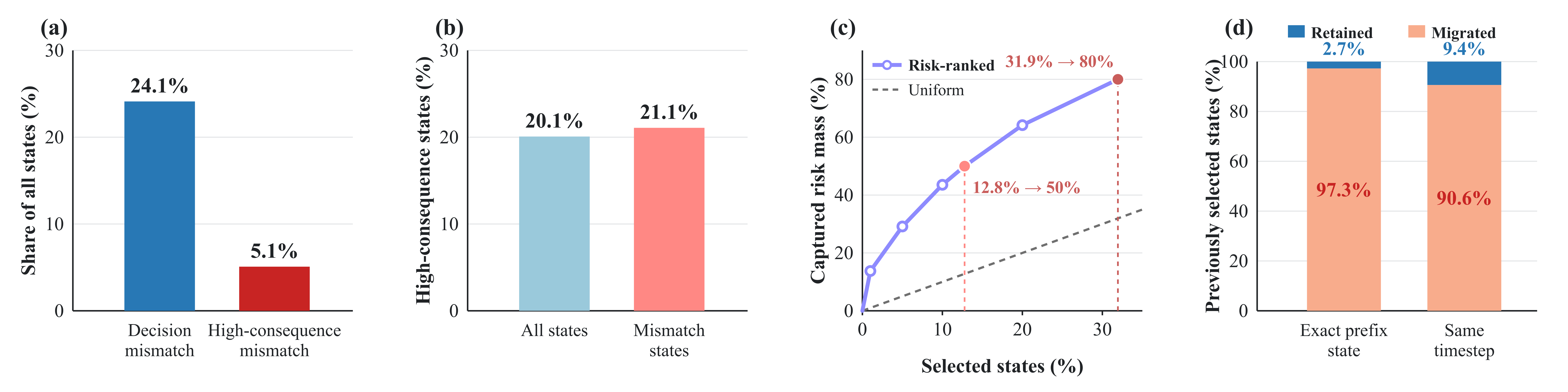}
    \caption{Properties of Decision--Consequence risk. (a) High-consequence
    mismatches are sparse. (b) Mismatch weakly enriches future consequence.
    (c) Risk mass concentrates in few states. (d) Selected states migrate after
    one \method{} update.}
    \label{fig:mismatch-consequence}
    \vspace{-0.6em}
\end{figure}

\noindent\textbf{Vision-Audio-Language MLLMs.}
Tab.~\ref{tab:omni-results} evaluates Qwen2.5-Omni-3B and
Qwen2.5-Omni-7B~\citep{xu2025qwen25omni} under W4A8 and W4A6. Across all four
settings, \method{} gives the lowest WER on both speech
benchmarks. Under W4A6, it also achieves the highest MMMU accuracy at both
model scales, while leading OmniBench in three of four combinations.

\subsection{Analysis}

\noindent\textbf{Decision mismatch is overinclusive.}
Fig.~\ref{fig:mismatch-consequence}(a) shows that top-1 mismatches cover
24.1\% of on-policy states, whereas states that are both mismatches and in the
top consequence quintile account for only 5.1\%. This gap motivates filtering
mismatches by future consequence.

\noindent\textbf{Mismatch weakly predicts future consequence.}
In Fig.~\ref{fig:mismatch-consequence}(b), high-consequence states constitute
20.1\% of all states and 21.1\% of mismatch states, indicating that current
token disagreement alone provides little information about future branch
divergence.

\noindent\textbf{Risk is concentrated in few states.}
Fig.~\ref{fig:mismatch-consequence}(c) shows that 12.8\% and 31.9\% of states
capture 50\% and 80\% of the total risk, respectively, supporting selective
risk weighting over uniform calibration.

\noindent\textbf{Selected states migrate after model updates.}
After one \method{} update, Fig.~\ref{fig:mismatch-consequence}(d) shows
retention rates of only 2.7\% under exact-prefix matching and 9.4\% under
prompt--timestep matching, motivating iterative trajectory refresh.

Together, these results support consequence-aware filtering, selective risk
weighting, and iterative on-policy refresh. Detailed protocols are provided in
Appendices~\ref{app:mismatch-consequence}, \ref{app:risk-concentration}, and
\ref{app:risk-migration}.

\begin{wraptable}{r}{0.40\textwidth}
    \centering
    \vspace{-0.6em}
    \caption{Fixed-budget selection on held-out joint-critical states.
    }
    \label{tab:selection-value}
    \small
    \setlength{\tabcolsep}{4pt}
    \resizebox{\linewidth}{!}{%
    \begin{tabular}{@{}lcc@{}}
        \toprule
        \textbf{Selection}
        & {\bfseries\shortstack{Mean($e_t$) $\downarrow$}}
        & {\bfseries\shortstack{$\Pr(m_Q^t<0)$ $\downarrow$}} \\
        \midrule
        Dense--InitQ                 & 2.322 & 69.23 \\
        Mismatch-only                & 2.446 & 61.54 \\
        \shortstack[l]{Mismatch +\\Near Miss}
                                     & \raisebox{1ex}{2.144}
                                     & \raisebox{1ex}{58.97} \\
        \rowcolor{onptqrow}
        \method{}                    & \textbf{1.609} & \textbf{53.85} \\
        \bottomrule
    \end{tabular}
    }
    \vspace{-0.6em}
\end{wraptable}

\noindent\textbf{Does counterfactual consequence improve state selection?}
We compare fixed-budget selectors from the same checkpoint while holding the
calibration data and optimization fixed. Mismatch-only uses top-1 disagreements;
Mismatch + Near Miss ranks both candidate types by $e_t+\eta j_t$; and \method{}
further incorporates counterfactual consequence $A_t$ through
Eq.~\ref{eq:risk}. The complete protocol is given in
Appendix~\ref{app:selection-value}.

Tab.~\ref{tab:selection-value} reports the mean boundary erosion
$\operatorname{Mean}(e_t)$ and the fraction of reference--competitor pairs whose
quantized margin flips sign, $\Pr(m_Q^t<0)$. Lower is better for both. \method{}
achieves the lowest values, showing that counterfactual consequence improves
state allocation beyond local decision signals.

\begin{wrapfigure}{r}{0.47\textwidth}
    \centering
  \vspace{-.6em}
    \includegraphics[width=\linewidth]{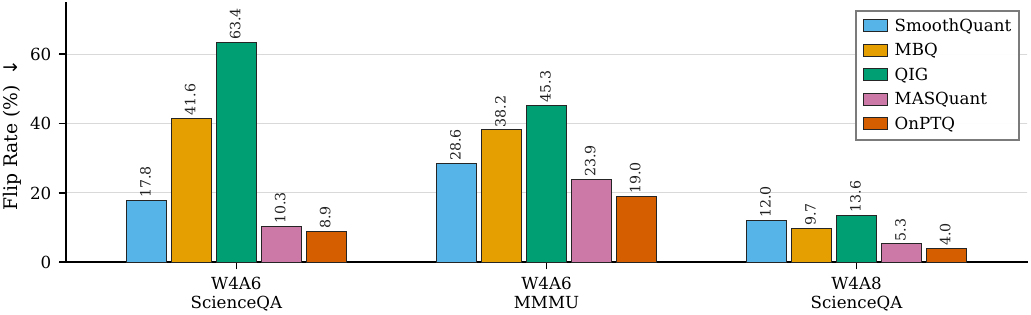}
    \caption{Sample-wise Flip Rate relative to the same Dense/FP16 reference on
    Qwen3-VL-8B-Instruct (lower is better). }
    \label{fig:flip-rate-grouped}
\end{wrapfigure}

\noindent\textbf{Behavior preservation beyond aggregate accuracy.}
Accuracy can obscure sample-level correctness changes when opposing
transitions cancel. We therefore report \emph{Flip Rate} in
Fig.~\ref{fig:flip-rate-grouped}, measuring the fraction of samples whose
correctness differs from the same Dense/FP16 reference. Across all three
evaluated settings, \method{} produces fewer correctness flips than the
compared PTQ methods, more closely preserving the Dense/FP16 model's
sample-wise correctness behavior.

\subsection{Ablation Studies}
\label{sec:ablations}

\begin{wraptable}{r}{0.45\textwidth}
    \centering
    \vspace{-0.6em}
    \caption{Core ablations on Qwen3-VL-8B-Instruct (W4A6); accuracy
    ($\uparrow$).}
    \label{tab:core-ablation}
    \scriptsize
    \setlength{\tabcolsep}{3pt}
    \renewcommand{\arraystretch}{0.96}
    \resizebox{\linewidth}{!}{%
    \begin{tabular}{@{}lcc@{}}
        \toprule
        \textbf{Setting}
        & \textbf{MMMU}
        & \textbf{ScienceQA} \\
        \midrule
        \multicolumn{3}{c}{\textbf{State Selection}} \\
        \midrule
        All on-policy states       & 45.4 & 86.8 \\
        Mismatch-only              & 47.1 & 88.3 \\
        \midrule
        \multicolumn{3}{c}{\textbf{Risk Composition}} \\
        \midrule
        Boundary only: $e_t$       & 44.9 & 86.5 \\
        w/o consequence $A_t$      & 45.7 & 87.3 \\
        w/o distribution term $j_t$& 47.5 & 88.6 \\
        KL instead of JS           & 48.0 & 89.0 \\
        \midrule
        \multicolumn{3}{c}{\textbf{Loss Composition}} \\
        \midrule
        w/o $\mathcal{L}_{\mathrm{anchor}}$   & 44.3 & 85.9 \\
        w/o $\mathcal{L}_{\mathrm{hidden}}$   & 46.6 & 87.8 \\
        w/o $\mathcal{L}_{\mathrm{pair}}$     & 45.5 & 87.1 \\
        w/o $\mathcal{L}_{\mathrm{response}}$ & 47.0 & 88.2 \\
        \midrule
        \multicolumn{3}{c}{\textbf{Policy Refresh}} \\
        \midrule
        Only one round             & 45.9 & 87.4 \\
        2 rounds, no refresh
                                   & 46.8 & 88.3 \\
                5 rounds, refresh
                                   & 44.2 & 86.1 \\
        \midrule
        \rowcolor{onptqrow}
        {\method{}}   & \textbf{48.8} & \textbf{89.7} \\
        \bottomrule
    \end{tabular}
    }
      \vspace{-1.4em}
\end{wraptable}

Tab.~\ref{tab:core-ablation} changes one design choice at a time under the same
W4A6 protocol. 

\noindent\textbf{State selection and risk composition.}
Using all on-policy states without candidate filtering performs worst among the
selection variants, while mismatch-only selection recovers part of the gap. The full
method further improves accuracy to 48.8 on MMMU and 89.7 on ScienceQA,
supporting the inclusion of near misses. Boundary erosion alone is insufficient:
removing $A_t$ costs 3.1 and 2.4 points, whereas removing $j_t$ costs 1.3 and
1.1 points. Thus, future consequence provides the larger gain, while the
same-state distribution term remains complementary. Replacing JS with KL is
competitive but remains below the full method.

\noindent\textbf{Loss composition and policy refresh.}
Removing any loss term lowers accuracy on both datasets. The context anchor has
the largest effect, followed by the pairwise boundary term, while hidden and
response reconstruction provide additional gains. One round reaches 45.9 and
87.4. A second round without trajectory refresh improves these values to 46.8
and 88.3, but remains below the full method with refreshed trajectories (48.8
and 89.7), isolating the benefit of recollecting states under the updated
quantized policy. Increasing the procedure to five refreshed rounds instead
reduces accuracy to 44.2 and 86.1, showing that more rounds are not
monotonically beneficial.

\subsection{Inference Efficiency}
\label{sec:inference-efficiency}

\begin{wraptable}{r}{0.38\textwidth}
    \centering
    \vspace{-0.8em}
    \caption{End-to-end prefill performance of Qwen2.5-VL-7B on a single
    NVIDIA B200 (sequence length $=2048$). BS: batch size.}
    \label{tab:inference-efficiency}
    \scriptsize
    \setlength{\tabcolsep}{4pt}
    \renewcommand{\arraystretch}{1.02}
    \resizebox{\linewidth}{!}{%
    \begin{tabular}{@{}lccc@{}}
        \toprule
        \textbf{Metric} & \textbf{BS} & \textbf{FP16} & \textbf{W4A8} \\
        \midrule
        Peak Memory (GB) & 1 & 16.22 & 10.15 \\
        Prefill (ms) &  1     & 36.73    & 20.48   \\
                \midrule
        Peak Memory (GB) & 8 & 21.13 & 15.06 \\
        Prefill (ms)  & 8     & 271.39    &  192.76   \\
        \bottomrule
    \end{tabular}
    }
    \vspace{-0.6em}
\end{wraptable}

Following the prefill protocol of MASQuant~\citep{hu2026masquant}, we compare
FP16 and W4A8 execution on a single NVIDIA B200 using the same model,
sequence length, and runtime configuration. As shown in
Tab.~\ref{tab:inference-efficiency}, W4A8 yields $1.79\times$ and $1.41\times$
prefill speedups at batch sizes 1 and 8, while reducing peak memory by 37.4\%
and 28.7\%, respectively. All additional computation introduced by \method{}---
on-policy trajectory collection, shared-state comparison, counterfactual
rollout, and risk scoring---is confined to offline calibration. At inference
time, only the calibrated W4A8 model is retained; neither the dense reference
nor any branching or risk-estimation module is executed. Consequently,
\method{} leaves the W4A8 inference graph and runtime cost unchanged while
retaining its efficiency gains over FP16.



\section{Conclusion}
\label{sec:conclusion}

This paper reframes post-training quantization for autoregressive MLLMs as an
on-policy calibration problem. \method{} uses shared-prefix comparison and
short counterfactual branches to prioritize decision-critical states, while
context anchoring and trajectory refresh maintain calibration coverage. Its
Decision--Consequence analysis links current policy discrepancy to future
consequence. Across vision--language and omni-modal Qwen models, \method{}
improves low-bit performance and reduces correctness flips without changing
the inference graph. Overall, autoregressive PTQ should prioritize errors that
alter future generation over numerical discrepancy alone.

\newpage
\section*{AI Assistance Disclosure}

Generative AI tools were used to draft selected passages, improve writing
clarity,formulating mathematical claims and retrieve and identify relevant literature. All AI-assisted text
and suggested references were reviewed, revised, and verified by the authors
against the original sources.  The authors take full responsibility for the final content
of this paper.

\bibliography{iclr2027_conference}
\bibliographystyle{iclr2027_conference}

\appendix
\clearpage
\noindent\textbf{Appendix organization.}
Appendix~\ref{app:algorithm} presents the complete \method{} calibration
algorithm, and Appendix~\ref{app:derivations} provides the theoretical
derivations. Appendix~\ref{app:experiments} details the implementation,
evaluation protocol, and hyperparameters, while
Appendix~\ref{app:supplementary-analyses} reports supplementary analyses that
support the motivating observations and state-selection design.
Appendix~\ref{app:limitations} discusses the limitations.
\section*{Reproducibility Statement}
The paper provides the complete calibration algorithm in
Appendix~\ref{app:algorithm}, the proof and surrogate correspondence in
Appendix~\ref{app:derivations}, and the calibration manifest, hyperparameters,
decoding protocol, and metric definitions in
Appendix~\ref{app:experiments}. Local comparisons use fixed prompts, decoding
settings, sample identities, and Dense/FP16 references.

\section{OnPTQ Algorithm}
\label{app:algorithm}

Algorithm~\ref{alg:onptq} summarizes the complete offline calibration procedure. The current quantized checkpoint generates every response prefix and every counterfactual continuation, while the frozen dense model supplies the reference action and evaluates branch separation. The routine updates the quantization calibration parameters $\theta$.

In round $k$, \textbf{Follow} first constructs the on-policy trajectory bank $\mathcal{B}^{(k)}$ defined in Sec.~\ref{sec:onptq-cycle}. \textbf{Compare} then extracts
\[
    \mathcal{C}_{\theta^{(k)}}
    =\left\{
      s_t\in\mathcal{B}^{(k)}
      \;\middle|\;
      s_t\text{ is a mismatch or near miss}
    \right\}.
\]
For each $s_t\in\mathcal{C}_{\theta^{(k)}}$, \textbf{Branch} stores the reference--competitor token pair and the resulting Decision--Consequence risk. The $\operatorname{TopK}$ step ranks these states by $r_t^{\mathrm{DC}}$ to obtain $\mathcal{S}_{\mathrm{critical}}^{(k)}$ for calibration.

\begin{algorithm}[H]
\caption{On-Policy Post-Training Quantization}
\label{alg:onptq}
\small
\begin{algorithmic}[1]
\Require Frozen dense model $D$; initial quantized model $Q_{\theta^{(0)}}$; unlabeled calibration set $\mathcal{D}_{\mathrm{cal}}$; rounds $R$; branch horizon $H$; critical-state budget $K_c$
\Ensure Calibrated quantized model $Q_{\theta^{(R)}}$
\For{$k=0,\ldots,R-1$}
    \State $\mathcal{B}^{(k)}\gets\varnothing$ \Comment{all on-policy states}
    \State $\mathcal{C}_{\theta^{(k)}}\gets\varnothing$ \Comment{candidate states}
    \For{each $(I,x)\in\mathcal{D}_{\mathrm{cal}}$}
        \State $y_{1:T}\gets\Call{GreedyRollout}{Q_{\theta^{(k)}},I,x}$
        \State Store the exact generated token IDs $y_{1:T}$
        \For{$t=1,\ldots,T$}
            \State $s_t\gets(I,x,y_{<t})$
            \State $\mathcal{B}^{(k)}\gets\mathcal{B}^{(k)}\cup\{s_t\}$
            \State Evaluate $z_D^t$ and $z_Q^t$ at the same state $s_t$
            \State $a_t^R\gets\arg\max_a z_D^t(a)$; $a_t^C\gets\arg\max_{a\neq a_t^R}z_Q^t(a)$
            \If{$s_t$ is a mismatch or near miss}
                \State Compute $e_t$ and $j_t$ using Equations~\ref{eq:erosion}--\ref{eq:same-state-js}
                \State $\tau_t^R\gets\Call{BranchRollout}{Q_{\theta^{(k)}},s_t,a_t^R,H}$
                \State $\tau_t^C\gets\Call{BranchRollout}{Q_{\theta^{(k)}},s_t,a_t^C,H}$
                \State Compute $A_t$ with frozen $D$ using Equation~\ref{eq:amplification}
                \State $r_t^{\mathrm{DC}}\gets e_t(1+\lambda_bA_t)+\eta j_t$
                \State $\mathcal{C}_{\theta^{(k)}}\gets\mathcal{C}_{\theta^{(k)}}\cup\{s_t\}$
                \State Store $(a_t^R,a_t^C,r_t^{\mathrm{DC}})$ for $s_t$
            \EndIf
        \EndFor
    \EndFor
    \State $\mathcal{S}_{\mathrm{critical}}^{(k)}\gets\operatorname{TopK}_{K_c}(\mathcal{C}_{\theta^{(k)}},r_t^{\mathrm{DC}})$
    \State Normalize $r_t^{\mathrm{DC}}$ over $\mathcal{S}_{\mathrm{critical}}^{(k)}$ to obtain $\widetilde w_t$
    \State Form $\mathcal{L}_{\mathrm{OnPTQ}}$ using Equations~\ref{eq:generic-objective}--\ref{eq:critical-loss}
    \State $\theta^{(k+1)}\gets\Call{BasePTQUpdate}{\theta^{(k)},\mathcal{L}_{\mathrm{OnPTQ}}}$
\EndFor
\State \Return $Q_{\theta^{(R)}}$
\end{algorithmic}
\end{algorithm}

\section{Additional Derivations}
\label{app:derivations}

\subsection{Proof of Theorem~1}

Fix a shared state $s$. Since $G(\tau)$ is bounded, $V_{\mu}(s,a)$ is finite for every $a\in\mathcal{A}$; since $\mathcal{A}$ is finite, the extrema
\begin{equation}
    V_{\max}(s)=\max_{a\in\mathcal{A}}V_{\mu}(s,a),
    \qquad
    V_{\min}(s)=\min_{a\in\mathcal{A}}V_{\mu}(s,a)
\end{equation}
exist. Define $\Delta\pi(a)=\pi_Q(a\mid s)-\pi_D(a\mid s)$. From the definition of the hybrid return,
\begin{equation}
    J_{Q\rightarrow\mu}(s)-J_{D\rightarrow\mu}(s)
    =\sum_{a\in\mathcal{A}}\Delta\pi(a)V_{\mu}(s,a).
    \label{eq:return-difference}
\end{equation}
Both policies are normalized, so $\sum_{a\in\mathcal{A}}\Delta\pi(a)=0$. Hence, for any constant $c$,
\begin{equation}
    \sum_{a\in\mathcal{A}}\Delta\pi(a)V_{\mu}(s,a)
    =\sum_{a\in\mathcal{A}}\Delta\pi(a)\big(V_{\mu}(s,a)-c\big).
    \label{eq:centered-return}
\end{equation}
Choose the midpoint $c=(V_{\max}(s)+V_{\min}(s))/2$. Then
\begin{equation}
    \left|V_{\mu}(s,a)-c\right|
    \leq \frac{V_{\max}(s)-V_{\min}(s)}{2}
    =\frac{\operatorname{span}_{a}V_{\mu}(s,\cdot)}{2}
\end{equation}
for every action. Applying the triangle inequality to Equation~\ref{eq:centered-return} gives
\begin{align}
    \left|J_{Q\rightarrow\mu}(s)-J_{D\rightarrow\mu}(s)\right|
    &\leq \sum_{a\in\mathcal{A}}|\Delta\pi(a)|\,
       \frac{\operatorname{span}_{a}V_{\mu}(s,\cdot)}{2} \\
    &=\TV\!\left(\pi_Q(\cdot\mid s),\pi_D(\cdot\mid s)\right)\,
       \operatorname{span}_{a}V_{\mu}(s,\cdot),
\end{align}
where the final equality uses
$\TV\!\left(\pi_Q(\cdot\mid s),\pi_D(\cdot\mid s)\right)=\frac{1}{2}\sum_{a\in\mathcal{A}}|\Delta\pi(a)|$.
This proves Equation~\ref{eq:dc-bound}. The argument also covers the degenerate cases $\pi_Q=\pi_D$ or $\operatorname{span}_{a}V_{\mu}(s,\cdot)=0$, for which both sides reduce appropriately. \hfill$\square$

\subsection{Pairwise Exact Decomposition}

Restrict the action set at state $s_t$ to the reference--competitor pair $\mathcal{P}_t=\{a_t^R,a_t^C\}$. For $M\in\{D,Q\}$, define the pair-normalized policy
\begin{equation}
    \widetilde\pi_M(a\mid s_t)
    =\frac{\pi_M(a\mid s_t)}
    {\pi_M(a_t^R\mid s_t)+\pi_M(a_t^C\mid s_t)},
    \qquad a\in\mathcal{P}_t.
    \label{eq:pair-policy}
\end{equation}
Let $q_M^t=\widetilde\pi_M(a_t^R\mid s_t)$. Since $\widetilde\pi_M(a_t^C\mid s_t)=1-q_M^t$, the pair-restricted return is
\begin{equation}
    J_{M\rightarrow\mu}^{\mathrm{pair}}(s_t)
    =q_M^t V_{\mu}(s_t,a_t^R)+(1-q_M^t)V_{\mu}(s_t,a_t^C).
\end{equation}
Subtracting the dense and quantized returns yields
\begin{equation}
    \left|J_{Q\rightarrow\mu}^{\mathrm{pair}}(s_t)-
            J_{D\rightarrow\mu}^{\mathrm{pair}}(s_t)\right|
    =\left|\widetilde\pi_Q(a_t^R\mid s_t)-
            \widetilde\pi_D(a_t^R\mid s_t)\right|
     \left|V_{\mu}(s_t,a_t^R)-V_{\mu}(s_t,a_t^C)\right|.
    \label{eq:pair-identity}
\end{equation}
Thus, on two actions, behavioral deviation decomposes exactly into shifted probability mass and an action-conditioned future-value difference.

For a softmax policy, the pairwise margin is also the pair-normalized log odds:
\begin{equation}
    m_M^t
    =z_M^t(a_t^R)-z_M^t(a_t^C)
    =\log\frac{\widetilde\pi_M(a_t^R\mid s_t)}
                    {\widetilde\pi_M(a_t^C\mid s_t)}.
    \label{eq:margin-log-odds}
\end{equation}
Consequently, $m_Q^t<m_D^t$ implies that quantization shifts pair-normalized probability mass from the reference action toward the competitor. The boundary erosion $e_t=[m_D^t-m_Q^t]_+$ measures the magnitude of this directional log-odds shift.

\subsection{From Theoretical Factors to OnPTQ Surrogates}

The theorem is expressed in terms of true action distributions and action-conditioned future values. \method{} instantiates these factors with label-free quantities available during calibration:
\begin{table}[H]
    \centering
    \small
    \caption{Correspondence between the Decision--Consequence analysis and the computable quantities used by \method{}.}
    \label{tab:theory-surrogates}
    \begin{tabular}{p{0.29\linewidth}p{0.63\linewidth}}
        \toprule
        Theoretical object & OnPTQ instantiation \\
        \midrule
        On-policy state $s_t\sim d_Q$ & Quant-policy trajectory collection \\
        Decision discrepancy & Boundary erosion $e_t$ and same-state JS $j_t$ \\
        Future consequence & Counterfactual branch amplification $A_t$ \\
        Policy-induced state shift & Iterative policy refresh \\
        \bottomrule
    \end{tabular}
\end{table}

Boundary erosion focuses on the selected reference--competitor pair, while $j_t$ detects changes elsewhere in the full next-token distribution. The inaccessible future-value difference in Equation~\ref{eq:pair-identity} is represented by $A_t$, which measures how far the two forced-token branches separate over a short quantized-model rollout. These quantities yield the practical risk in Equation~\ref{eq:risk}; the theorem supplies its decision--consequence structure, while the surrogate quality is evaluated empirically.

The statewise result becomes an on-policy objective by taking expectation over the states induced by the quantized model:
\begin{equation}
    \mathcal{R}_{\mathrm{on}}(Q)
    =\mathbb{E}_{s_t\sim d_Q}
      \left[
      \TV\!\left(\pi_Q(\cdot\mid s_t),\pi_D(\cdot\mid s_t)\right)\,
      \operatorname{span}_{a}V_{\mu}(s_t,\cdot)
      \right].
    \label{eq:on-policy-risk}
\end{equation}
Static PTQ changes both factors in Equation~\ref{eq:on-policy-risk} only indirectly and evaluates them under a different state distribution. \method{} instead samples $d_Q$ explicitly and constructs separate surrogates for the decision and consequence terms.

\section{Experimental Setup and Hyperparameters}
\label{app:experiments}

\subsection{Implementation and evaluation protocol}
\label{app:implementation-details}

\paragraph{Hardware.}
All experiments were conducted on a server equipped with eight NVIDIA B200
GPUs. The inference-efficiency measurements in
Sec.~\ref{sec:inference-efficiency} use a single B200, as specified there.

\paragraph{Quantization scope.}
Only the language-model component is quantized; the visual and audio encoders
remain at their original precision. W4A16 denotes 4-bit weights with 16-bit
activations, whereas W8A8, W4A8, and W4A6 jointly quantize weights and
activations. We use per-channel weight quantization and dynamic per-token
activation quantization.

\paragraph{Result provenance.}
For Qwen2.5-VL-7B, the available Dense and baseline results in
Tab.~\ref{tab:main-results} are transcribed from the original MASQuant paper;
the \method{} rows are evaluated separately under the settings reported here.
For Qwen3-VL-8B-Instruct, we reproduce the compared methods from their released
open-source implementations and evaluate them with a common local pipeline.

\paragraph{Calibration datasets.}
Following QIG~\citep{xiang2026qig}, all matched vision--language runs use the
same frozen 128-pair manifest. We sample uniformly without replacement from
50,027 ShareGPT4V-improved COCO Caption records using seed 26260725. The COCO
image ID is the stable sample identifier, whereas the ordinal denotes its
zero-based position in the frozen manifest. The complete manifest is listed
below; each pair is formatted with the conversational template of the target
model (Tab.~\ref{tab:calibration-ids}).
\begin{table}[H]
\centering
\caption{COCO image IDs of the frozen 128-sample vision--language calibration
manifest. Ord. denotes the zero-based manifest order.}
\label{tab:calibration-ids}
\scriptsize
\setlength{\tabcolsep}{2.5pt}
\renewcommand{\arraystretch}{0.92}
\begin{tabular}{@{}r l r l r l r l@{}}
\toprule
\textbf{Ord.} & \textbf{COCO image ID} & \textbf{Ord.} & \textbf{COCO image ID}
& \textbf{Ord.} & \textbf{COCO image ID} & \textbf{Ord.} & \textbf{COCO image ID} \\
\midrule
0 & \texttt{000000039043} & 32 & \texttt{000000127647} & 64 & \texttt{000000222468} & 96 & \texttt{000000069410} \\
1 & \texttt{000000153506} & 33 & \texttt{000000201420} & 65 & \texttt{000000044928} & 97 & \texttt{000000170040} \\
2 & \texttt{000000098431} & 34 & \texttt{000000071044} & 66 & \texttt{000000117289} & 98 & \texttt{000000007232} \\
3 & \texttt{000000130270} & 35 & \texttt{000000100034} & 67 & \texttt{000000235642} & 99 & \texttt{000000091912} \\
4 & \texttt{000000016706} & 36 & \texttt{000000187474} & 68 & \texttt{000000211041} & 100 & \texttt{000000034785} \\
5 & \texttt{000000173142} & 37 & \texttt{000000207597} & 69 & \texttt{000000066866} & 101 & \texttt{000000108531} \\
6 & \texttt{000000082142} & 38 & \texttt{000000215303} & 70 & \texttt{000000230177} & 102 & \texttt{000000022051} \\
7 & \texttt{000000073141} & 39 & \texttt{000000042743} & 71 & \texttt{000000245733} & 103 & \texttt{000000103331} \\
8 & \texttt{000000009807} & 40 & \texttt{000000088286} & 72 & \texttt{000000211919} & 104 & \texttt{000000057801} \\
9 & \texttt{000000038685} & 41 & \texttt{000000003737} & 73 & \texttt{000000167264} & 105 & \texttt{000000068120} \\
10 & \texttt{000000146570} & 42 & \texttt{000000028540} & 74 & \texttt{000000239532} & 106 & \texttt{000000032523} \\
11 & \texttt{000000076590} & 43 & \texttt{000000220097} & 75 & \texttt{000000140073} & 107 & \texttt{000000240033} \\
12 & \texttt{000000110547} & 44 & \texttt{000000239499} & 76 & \texttt{000000120241} & 108 & \texttt{000000154354} \\
13 & \texttt{000000227540} & 45 & \texttt{000000062307} & 77 & \texttt{000000056233} & 109 & \texttt{000000200483} \\
14 & \texttt{000000227802} & 46 & \texttt{000000031747} & 78 & \texttt{000000220182} & 110 & \texttt{000000105688} \\
15 & \texttt{000000166320} & 47 & \texttt{000000038902} & 79 & \texttt{000000164042} & 111 & \texttt{000000130287} \\
16 & \texttt{000000160103} & 48 & \texttt{000000109278} & 80 & \texttt{000000048491} & 112 & \texttt{000000049819} \\
17 & \texttt{000000172467} & 49 & \texttt{000000068359} & 81 & \texttt{000000189957} & 113 & \texttt{000000146504} \\
18 & \texttt{000000218772} & 50 & \texttt{000000152823} & 82 & \texttt{000000116088} & 114 & \texttt{000000201632} \\
19 & \texttt{000000194034} & 51 & \texttt{000000242734} & 83 & \texttt{000000215708} & 115 & \texttt{000000202387} \\
20 & \texttt{000000176414} & 52 & \texttt{000000190172} & 84 & \texttt{000000041818} & 116 & \texttt{000000095595} \\
21 & \texttt{000000047935} & 53 & \texttt{000000149588} & 85 & \texttt{000000158726} & 117 & \texttt{000000077750} \\
22 & \texttt{000000014766} & 54 & \texttt{000000025060} & 86 & \texttt{000000043555} & 118 & \texttt{000000035299} \\
23 & \texttt{000000059540} & 55 & \texttt{000000049363} & 87 & \texttt{000000028097} & 119 & \texttt{000000119979} \\
24 & \texttt{000000221346} & 56 & \texttt{000000195188} & 88 & \texttt{000000170211} & 120 & \texttt{000000103722} \\
25 & \texttt{000000105804} & 57 & \texttt{000000118302} & 89 & \texttt{000000000599} & 121 & \texttt{000000046936} \\
26 & \texttt{000000205473} & 58 & \texttt{000000125661} & 90 & \texttt{000000063307} & 122 & \texttt{000000204805} \\
27 & \texttt{000000187866} & 59 & \texttt{000000112577} & 91 & \texttt{000000081906} & 123 & \texttt{000000217376} \\
28 & \texttt{000000229852} & 60 & \texttt{000000208886} & 92 & \texttt{000000052386} & 124 & \texttt{000000206469} \\
29 & \texttt{000000090683} & 61 & \texttt{000000094544} & 93 & \texttt{000000186606} & 125 & \texttt{000000067814} \\
30 & \texttt{000000134375} & 62 & \texttt{000000209925} & 94 & \texttt{000000105899} & 126 & \texttt{000000099030} \\
31 & \texttt{000000113722} & 63 & \texttt{000000115912} & 95 & \texttt{000000103267} & 127 & \texttt{000000179793} \\
\bottomrule
\end{tabular}
\end{table}
For the matched Qwen2.5-Omni-7B runs, the frozen 128-example calibration
manifest contains OmniBench IDs 0--127.

\paragraph{OnPTQ calibration.}
All matched runs start from the same initialized quantized checkpoint, and
\method{} updates only the quantization calibration parameters $\theta$. Unless
otherwise stated, we use greedy on-policy rollouts with at most 64 response
tokens, two rounds with trajectory refresh, branch horizon $H=4$, critical-state
budget $K_c=256$, $\lambda_b=\eta=1.0$, and $\gamma=0.2$. The full procedure is given in
Algorithm~\ref{alg:onptq}, and the sensitivity of these choices is reported in
Appendix~\ref{app:hyperparameters}.

\paragraph{Evaluation protocol.}
Methods within each matched comparison use identical task prompts, decoding,
and evaluation samples. The local Qwen2.5-Omni-7B evaluation uses 900 MMMU
validation examples. Tab.~\ref{tab:omni-results} reports the 1,014-example
OmniBench evaluation split after excluding calibration IDs 0--127. Accuracy is
reported for MMMU, OCRBench, VizWiz,
ScienceQA, TextVQA, and OmniBench; lower word error rate is better for
LibriSpeech and WenetSpeech. The paired Accuracy--Flip evaluation is restricted
to MMMU and ScienceQA-IMG and compares each quantized prediction with the same
Dense/FP16 prediction by sample identity; Appendix~\ref{app:accuracy-flip}
gives the complete definition.

\subsection{Hyperparameter Analysis}
\label{app:hyperparameters}

Tab.~\ref{tab:onptq-hyperparameters} varies one method-specific hyperparameter
at a time while fixing the others to the main configuration. Here, $H$ is the
branch horizon, $K_c$ is the number of critical states retained in each round,
$\lambda_b$ and $\eta$ balance the two risk terms, and $\gamma$ weights
critical-state calibration.
The number of on-policy rounds $R$ is studied separately in
Tab.~\ref{tab:core-ablation}. All variants use the same calibration data,
initialized quantized model, and optimization schedule; only the listed
hyperparameter changes.

In the risk-composition ablation, \emph{KL instead of JS} replaces the JS
divergence in both the same-state term $j_t$ and the per-step branch divergence
used to compute $A_t$; the remaining risk and optimization settings are fixed.
In the policy-refresh ablation, \emph{2 rounds, no refresh} reuses the first
round's trajectory bank for the second update, whereas \emph{5 rounds,
refresh} recollects the trajectory bank after every model update.

\begin{table}[H]
    \centering
    \caption{One-factor-at-a-time hyperparameter analysis on
    Qwen3-VL-8B-Instruct under W4A6. Shaded rows denote the main settings; all
    values are accuracy ($\uparrow$).}
    \label{tab:onptq-hyperparameters}
    \small
    \setlength{\tabcolsep}{12pt}
    \renewcommand{\arraystretch}{0.98}
    \begin{tabular}{lcc}
        \toprule
        \textbf{Setting} & \textbf{MMMU} & \textbf{ScienceQA} \\
        \midrule
        \multicolumn{3}{c}{\textbf{Branch horizon $H$}} \\
        \midrule
        2 & 47.9 & 88.9 \\
        \rowcolor{onptqrow}
        \textbf{4} & 48.8 & 89.7 \\
        8 & 48.3 & 89.2 \\
        \midrule
        \multicolumn{3}{c}{\textbf{Critical-state budget $K_c$}} \\
        \midrule
        64 & 47.4 & 88.5 \\
        128 & 48.1 & 89.1 \\
        \rowcolor{onptqrow}
        \textbf{256} & 48.8 & 89.7 \\
        \midrule
        \multicolumn{3}{c}{\textbf{Branch weight $\lambda_b$}} \\
        \midrule
        0.5 & 47.9 & 88.9 \\
        \rowcolor{onptqrow}
        \textbf{1.0} & 48.8 & 89.7 \\
        2.0 & 48.2 & 89.1 \\
        \midrule
        \multicolumn{3}{c}{\textbf{Distribution weight $\eta$}} \\
        \midrule
        0.5 & 48.3 & 89.2 \\
        \rowcolor{onptqrow}
        \textbf{1.0} & 48.8 & 89.7 \\
        2.0 & 48.1 & 89.0 \\
        \midrule
        \multicolumn{3}{c}{\textbf{Critical-loss weight $\gamma$}} \\
        \midrule
        0.1 & 47.8 & 88.8 \\
        \rowcolor{onptqrow}
        \textbf{0.2} & 48.8 & 89.7 \\
        0.4 & 48.0 & 89.0 \\
        \bottomrule
    \end{tabular}
\end{table}

The two datasets exhibit consistent trends. Increasing $K_c$ from 64 to 256
steadily improves accuracy, showing the benefit of covering more critical
states within the evaluated budget. The branch horizon peaks at $H=4$; extending
it to eight steps provides no additional gain under this protocol. Both risk
coefficients perform best at $\lambda_b=\eta=1.0$, while $\gamma=0.2$ gives the
best balance between critical-state calibration and the context anchor. Across
all sweeps, non-default settings remain within 1.4 points on MMMU and 1.2 points
on ScienceQA of the main configuration, indicating moderate sensitivity without
performance collapse.

\subsection{Accuracy--Flip evaluation}
\label{app:accuracy-flip}

For the closed-choice ScienceQA-IMG and MMMU evaluations, every quantized
prediction is paired by sample identity with the same Dense/FP16 prediction.
Let $c_i^D,c_i^Q\in\{0,1\}$ denote their correctness indicators. Flip Rate is
$N^{-1}\sum_i\mathbf{1}\{c_i^D\neq c_i^Q\}$, equivalently the sum of
correct-to-incorrect and incorrect-to-correct transition rates. In contrast,
the accuracy change is their difference, so opposing transitions can cancel.
Fig.~\ref{fig:flip-rate-grouped} reports single-run paired estimates on
ScienceQA-IMG ($N{=}2{,}017$) and MMMU ($N{=}900$), using the same evaluation
samples and Dense/FP16 reference for every method.

\section{Supplementary Analyses}
\label{app:supplementary-analyses}

The mismatch--consequence, risk-concentration, and risk-migration analyses in
Secs.~\ref{app:mismatch-consequence}--\ref{app:risk-migration} are computed from
model trajectories without downstream labels or task scores.

\subsection{Motivating diagnostics}

Sec.~\ref{sec:motivation} uses Qwen2.5-VL-7B-Instruct with W4A8 QIG and MASQuant checkpoints; their calibration settings follow Appendix~\ref{app:implementation-details}. The state-coverage and fork-aligned diagnostics use a 64-example state bank and 120 non-overlapping ScienceQA-IMG examples for evaluation. Rollouts use greedy decoding with at most 64 response tokens. Counterfactual branches follow the current quantized policy with horizon $H=4$; if either branch terminates early, statistics are computed over the common valid horizon.

For the anchored on-policy diagnostic in Observation III, the closed-loop JS change relative to full-sequence static calibration is $-0.0125$ with a paired 95\% interval of $[-0.0241,-0.0007]$; the corresponding accuracy change is $+0.005$ with interval $[-0.010,0.020]$. Across candidate states, local JS and counterfactual branch amplification have Pearson correlation $0.232$, Spearman correlation $0.170$, and top-20\% overlap $21.15\%$.

\subsection{Mismatch--consequence analysis}
\label{app:mismatch-consequence}

Fig.~\ref{fig:mismatch-consequence} uses Qwen2.5-VL-7B-Instruct under W4A8
quantization and 32 source-disjoint held-out prompts, yielding 767 states with
valid $H{=}16$ counterfactual branches. A state is a decision mismatch when the
dense and quantized top-1 tokens differ, and is high consequence when its
$H{=}16$ branch divergence lies in the top quintile. Uncertainty is estimated
with 2,000 prompt-cluster bootstrap replicates.

\subsection{Risk-concentration analysis}
\label{app:risk-concentration}

Fig.~\ref{fig:mismatch-consequence}(c) uses the same Qwen2.5-VL-7B-Instruct W4A8
checkpoint and 767 held-out on-policy states as the mismatch--consequence
analysis. We rank states by the full Decision--Consequence risk and compute the
fraction of its total mass captured by each selection budget. The top 1\%,
5\%, 10\%, and 20\% of states capture 13.8\%, 29.1\%, 43.5\%, and 64.2\% of
the total risk, respectively.

\subsection{Risk-migration analysis}
\label{app:risk-migration}

Fig.~\ref{fig:mismatch-consequence}(d) compares the states selected on the same
128-prompt calibration manifest before and after the first \method{} update.
Each round stores two selected states per prompt, yielding 256 states. Exact
retention requires both the prompt identity and quantized prefix token sequence
to match; timestep retention requires only the prompt identity and generation
timestep to match. Both rounds use the same rollout seed schedule. Because the
stored rollouts contain only the selected states rather than every token-level
risk value, we report migration through selected-state retention rather than
changes in the complete risk ranking.

\subsection{State-selection value analysis}
\label{app:selection-value}

\paragraph{Setup and controlled variants.}
Tab.~\ref{tab:selection-value} uses Qwen2.5-VL-7B-Instruct under W4A8. All
calibrated variants start from the same Dense--InitQ checkpoint and use the same
128 calibration prompts. Each prompt contributes one mismatch and one near-miss
candidate, from which each rule selects one state, giving an identical budget of
128 states. State weights are fixed to one, and the optimized parameters, loss
terms, optimizer, and update schedule are shared; only the selection rule
changes. Dense--InitQ receives no additional update. Mismatch-only selects the
mismatch; Mismatch + Near Miss selects the higher-ranked candidate under
$e_t+\eta j_t$; Consequence-only uses $A_t$; \method{} uses
$r_t^{\mathrm{DC}}$; and Random control samples from the same two-candidate pool.

\paragraph{Held-out evaluation.}
We freeze a common evaluation bank before comparing checkpoints: 32
source-disjoint prompts yield 767 valid states, including 185 dense--quantized
top-1 mismatches and 39 joint-critical states. A joint-critical state is both a
top-1 mismatch and in the top quintile of independently measured $H{=}16$
branch consequence. Every checkpoint is evaluated on identical quantized
prefixes and identical reference--competitor action pairs.
Tab.~\ref{tab:selection-value} reports
the joint-critical subset; Tab.~\ref{tab:selection-value-full} additionally
reports all states and the full mismatch subset.

\paragraph{Metrics and uncertainty.}
For a frozen state set $\mathcal{S}$, we report
\begin{equation}
    \overline e_{\mathcal{S}}
    =\frac{1}{|\mathcal{S}|}\sum_{t\in\mathcal{S}}[m_D^t-m_Q^t]_+,
    \qquad
    \operatorname{Flip}_{\mathcal{S}}
    =\frac{100}{|\mathcal{S}|}\sum_{t\in\mathcal{S}}
      \mathbf{1}\{m_Q^t<0\}.
    \label{eq:selection-metrics}
\end{equation}
The first metric measures how much quantization erodes the dense
reference--competitor margin; the second measures how often that erosion reverses
the pair ordering. Lower values mean that the quantized checkpoint more closely
preserves the dense decision boundary. Pairwise uncertainty is estimated with a
paired prompt-cluster bootstrap, treating prompts rather than individual tokens
as independent sampling units.

\begin{table}[H]
    \caption{Complete fixed-state recovery results for alternative
    state-selection rules (lower is better).}
    \label{tab:selection-value-full}
    \centering
    \small
    \begin{tabular}{@{}lcccccc@{}}
        \toprule
        \textbf{Selection}
        & {\bfseries\shortstack{All\\erosion}}
        & {\bfseries\shortstack{All\\flip (\%)}}
        & {\bfseries\shortstack{Mismatch\\erosion}}
        & {\bfseries\shortstack{Mismatch\\flip (\%)}}
        & {\bfseries\shortstack{Joint-critical\\erosion}}
        & {\bfseries\shortstack{Joint-critical\\flip (\%)}} \\
        \midrule
        Dense--InitQ       & 1.608 & 22.43 & 2.600 & 64.86 & 2.322 & 69.23 \\
        Mismatch-only      & 1.303 & 19.95 & 2.363 & 50.81 & 2.446 & 61.54 \\
        Mismatch + Near Miss
                           & \textbf{1.032} & 17.73 & 2.248 & 52.43 & 2.144 & 58.97 \\
        Consequence-only   & 1.073 & \textbf{15.91} & 2.003 & 45.95 & 1.664 & \textbf{43.59} \\
        \rowcolor{onptqrow}
        \method{}          & 1.069 & 19.04 & \textbf{1.642} & \textbf{44.86}
                           & \textbf{1.609} & 53.85 \\
        Random control     & 1.216 & 20.99 & 2.407 & 55.68 & 2.199 & 64.10 \\
        \bottomrule
    \end{tabular}
\end{table}

This controlled experiment measures fixed-state decision-boundary recovery;
downstream task performance is evaluated separately in Sec.~\ref{sec:experiments}.

\subsection{Single-signal diagnostics}
\label{app:single-signal-diagnostics}

Tab.~\ref{tab:single-signal-diagnostics} compares how each state-level signal
relates to current mismatch and future consequence. Same-state JS and boundary
erosion primarily track current mismatch, whereas branch consequence is more
strongly aligned with future consequence. The full \method{} risk integrates
these complementary signals for state selection.

\begin{table}[H]
    \caption{Diagnostics of individual signals and the \method{} risk for
    current mismatch and future consequence.}
    \label{tab:single-signal-diagnostics}
    \centering
    \small
    \begin{tabular}{@{}lcc@{}}
        \toprule
        \textbf{Signal}
        & {\bfseries\shortstack{Mismatch\\AUROC $\uparrow$}}
        & {\bfseries\shortstack{Future\\Spearman $\uparrow$}} \\
        \midrule
        Hidden reconstruction error & 0.470 & -0.050 \\
        Same-state JS divergence     & \textbf{0.870} & 0.111 \\
        Boundary erosion $e_t$       & 0.821 & 0.077 \\
        Branch consequence $A_t$     & 0.543 & \textbf{0.504} \\
        \rowcolor{onptqrow}
        \method{} risk $r_t^{\mathrm{DC}}$ & 0.822 & 0.127 \\
        \bottomrule
    \end{tabular}
\end{table}

\section{Limitations}
\label{app:limitations}

\method{} adds offline calibration cost through dense--quantized comparisons
and short greedy rollouts, whose local consequence surrogate may miss
longer-horizon or stochastic effects. Evaluation is currently limited to
Qwen-based multimodal models.

\end{document}

%% file: math_commands.tex
\usepackage{amsmath,amsfonts,bm}

\def\eqref#1{equation~\ref{#1}}

\def\1{\bm{1}}

\DeclareMathAlphabet{\mathsfit}{\encodingdefault}{\sfdefault}{m}{sl}
\SetMathAlphabet{\mathsfit}{bold}{\encodingdefault}{\sfdefault}{bx}{n}

